\documentclass{article}

\usepackage{microtype}
\usepackage{graphicx}
\usepackage[font=small,labelfont=bf]{caption} 

\usepackage{subfigure}
\usepackage{booktabs} 
\usepackage{amsmath}
\usepackage{mathtools}
\usepackage{amssymb}
\usepackage{siunitx} 
\usepackage{xspace}
\usepackage{comment}
\usepackage{svg}
\usepackage{float}
\usepackage{textcomp}
\usepackage[section]{placeins} 
\usepackage{dblfloatfix} 

\usepackage[sort&compress]{natbib}  

\usepackage{hyperref}

\usepackage[accepted]{mlsys2025}

\usepackage{soul}        
\usepackage{xcolor}

\mlsystitlerunning{DiffPro: Joint Timestep and Layer Wise Precision Optimization for Efficient Diffusion Inference}
\begin{document}




\twocolumn[
\mlsystitle{DiffPro: Joint Timestep and Layer-Wise Precision Optimization for Efficient Diffusion Inference}




\mlsyssetsymbol{equal}{*}

\begin{mlsysauthorlist}
\mlsysauthor{Farhana Amin}{to}
\mlsysauthor{Sabiha Afroz}{to}
\mlsysauthor{Kanchon Gharami}{goo}
\mlsysauthor{Mona Moghadampanah}{to}
\mlsysauthor{Dimitrios S. Nikolopoulos}{to}
\end{mlsysauthorlist}

\mlsysaffiliation{to}{Department of Comuputer Science, Virginia Tech, Blacksburg, Virginia, USA}
\mlsysaffiliation{goo}{Department of Electrical Engineering \& Comuputer Science, Embry Riddle Aeronautical University, Daytona Beach, Florida, USA}

\mlsyscorrespondingauthor{Farhana Amin}{afarhana@vt.edu}




\mlsyskeywords{Machine Learning, MLSys}

\vskip 0.3in

\begin{abstract}
Diffusion models produce high quality images but inference is costly due to many denoising steps and heavy matrix operations. We present DiffPro, a post training, hardware faithful framework that works with the exact integer kernels used in deployment and jointly tunes timesteps and per layer precision in Diffusion Transformers(DiTs) to reduce latency and memory without any training. DiffPro combines three parts: a manifold aware sensitivity metric to allocate weight bits, dynamic activation quantization to stabilize activations across timesteps, and a budgeted timestep selector guided by teacher–student drift. In experiments DiffPro achieves up to $6.25{\times}$ model compression, fifty percent fewer timesteps, and $2.8{\times}$ faster inference with $\Delta \mathrm{FID} \leq 10$ on standard benchmarks, demonstrating practical efficiency gains. DiffPro unifies step reduction and precision planning into a single budgeted deployable plan for real time energy aware diffusion inference.
\end{abstract}

]



\printAffiliationsAndNotice{}
\section{Introduction}
\label{sec:intro}
Diffusion models have brought a new paradigm for high-fidelity image and video generation. Their ability to model complex data distributions through iterative denoising has surpassed Generative Adversarial Networks and autoregressive approaches in visual quality and scalability. However, this quality comes at a significant cost. Hundreds of inference steps and large matrix multiplications across many blocks make diffusion models computationally expensive to deploy. While training can be distributed across large clusters, per request inference directly limits latency, throughput, memory, and energy efficiency~\citep[p.~28312]{chen2025}~\cite{sun2024tmpq}~\citep[p.~152] {ye2025pqd}.

Recent research has explored several ways to speed up diffusion inference, such as reducing denoising steps, using latent space formulations, and applying quantization to reduce memory and computation. However, these approaches typically address the problem in isolation. Timestep reduction speeds up inference but often damages image quality when too many steps are removed. Quantization improves efficiency but can be unstable, especially in Diffusion Transformers (DiTs)~\citep[p.~4196]{peebles2023scalable}. Quantizing DiTs~\cite{bao2023one} is particularly challenging because they exhibit a strong temporal drift in activation ranges across timesteps, making static calibration unreliable. They also suffer from channel imbalance and time dependent variation, where certain channels or layers are more sensitive to low bit operations. These effects make low precision inference fragile unless scaling is carefully adapted. The real challenge lies in this uneven behavior, 
DiTs~\cite{bao2023one} behave differently at each timestep and across layers. Using the same lower bit for all layers or a fixed sampling schedule ignores these variations, causing gradual quality loss through the diffusion process. Hence, a single, ideal solution is still missing that can balance precision, image quality, and computational cost together~\citep[p.~16027]{wang2024towards}.

In this paper, we present DiffPro, a post training optimization framework which jointly tunes layer wise bit precision and timestep budgets for efficient diffusion inference. We propose a hardware faithful optimization framework where we measure accuracy–latency using integer kernels and quantization operations(op) that run in deployment time(e.g., INT8×INT8→INT32 GEMMs). DiffPro swaps linear layers for those kernels, packs weights with GPTQ~\cite{frantar2022gptq}, applies runtime Dynamic Activation Quantization (DAQ) before each op, and profiles with CUDA timers so results match deployed compute and memory layouts. DiffPro introduces three key mechanisms: (i) manifold aware sensitivity estimation that ranks layers via principal component analysis(PCA) and curvature energy to assign bits where they matter most (ii) DAQ that adjusts activation scales per sample and timestep to mitigate temporal drift and (iii) budgeted timestep selection that prunes redundant denoising steps based on teacher-student drift analysis. A final joint plan search integrates these components into a deployable configuration optimized for both mean squared accuracy and latency under real integer kernels. Our key contributions are summarized as follows:

\begin{itemize}
    \item We present DiffPro, the first post training framework that jointly optimizes diffusion timesteps and per layer quantization precision under one compute budget.
    
    \item We design a manifold aware sensitivity metric linking layer curvature energy and activation variance to quantization robustness, providing a geometric basis for precision allocation in Diffusion Transformers.
    
    \item We introduce DAQ to mitigate timestep dependent activation drift, dynamically rescaling activations for stable low bit inference across the denoising process.
    
    \item We propose a budgeted timestep selection method guided by teacher-student drift, removing redundant steps while maintaining late stage fidelity through informed pruning. DiffPro delivers up to 6.25$\times$ model compression, 50\% timestep reduction with only 500 timesteps and 2.8$\times$ faster inference with $\Delta\mathrm{FID}$
    $\leq 10$ on CIFAR-10 \& CIFAR-100~\cite{krizhevsky_cifar_2009}, Imagenet~\cite{deng2009imagenet} and CelebA-HQ~\cite{karras2018celebAHQ} datasets, achieving hardware faithful efficiency without retraining.
\end{itemize}

This work differs with prior DiT~\citet{bao2023one} post training quantization~\cite{sun2024tmpq} by choosing the number of sampling steps and each layer’s precision together, under a latency budget measured on real hardware. Guided by teacher–student drift and a manifold aware sensitivity score, we get faster sampling without hurting quality.
\section{Background and Motivation}
\label{sec:background work}

Diffusion models denoise a noise sample using a timestep-conditioned backbone such as U-Net~\citet{ronneberger2015u} or DiT~\citet{bao2023one}. Similar computations are repeated over hundreds of steps. Common speedups like quantization with fewer bits and sampling with fewer steps often use uniform bit widths and uniform schedules~\cite{shang2023post}, overlooking per layer and per timestep sensitivity and degrading quality.


\noindent Existing diffusion compression uses uniform quantization despite activation variation across timesteps, prunes steps uniformly despite unequal contributions, and lacks diffusion aware dynamic analysis of layer and timestep importance~\cite{shang2023post}. These choices waste compute and reduce fidelity. Prior work~\cite{sun2024tmpq} does not jointly choose per layer bit widths and a reduced timestep set under one compute or memory budget with sensitivity signals tied to the deployed quantized model. This motivates a sensitivity aware policy that jointly optimizes precision and sampling.


\subsection{From U-Nets to DiTs and the Cost of Sampling}
Diffusion models denoise over many timesteps, so latency scales with step count. DiTs~\citet{bao2023one} replace U-Nets~\citet{ronneberger2015u} for scalability and quality, yet their activations change over time rather than monotonically shrinking. Prior work assumes static activation scales or fixes the step count, missing DiT~\citet{bao2023one} specific temporal behavior~\cite{vora2025ptq4adm,FP4DiT2025}. We therefore model timestep statistics explicitly and optimize the number of steps where it most reduces latency.


\subsection{Quantization and Timestep}
Quantization lowers precision to save compute, and sampling reduces the number of steps; both shape the latency–quality balance. DiTs~\citet{abouelnaga2016cifar} vary across space and time, so uniform bit widths and uniform schedules can hurt quality. Prior work pushes either quantization or scheduling, often tuning them separately and without hardware faithful checks~\cite{chen2025,ViDiTQ2025,FP4DiT2025,TRDQ2025,HTGDiT2025}. We treat them together and measure with real $ \mathrm{INT8}\times\mathrm{INT8}\rightarrow\mathrm{INT32} $ kernels. Not every denoising step pulls its weight. We prune for speed and keep a short late step tail for stability. Earlier studies often fix or thin steps uniformly and do not pick steps and precision under one shared budget~\cite{HTGDiT2025}. We track teacher–student drift, protect the tail, choose the top $K$ steps under a compute or memory limit, and set activation bits and weight plans together for end to end gains.


\subsection{DAQ and Temporal Drift}
Activation distributions shift along the diffusion trajectory, so per timestep scaling is needed. Dynamic or grouped online scales handle timestep dependent outliers better than static tensor wise scales, yet prior work largely underuses this capability~\cite{chen2025,vora2025ptq4adm,ViDiTQ2025,FP4DiT2025}. We add lightweight prehooks that compute per sample per timestep min max scales calibrated to student drift with minimal overhead.


\subsection{Layer Sensitivity Using Manifold Structure}
Layers differ in sensitivity, so bit budgets must be layer wise. Magnitude signals such as Hessian diagonal input sensitivity$\sum x^2$ and manifold signals such as PCA rank capture complementary behavior. Prior DiT~\citet{bao2023one} quantization often uses one signal or fixed mixed precision rules, which misallocate bits as statistics change~\cite{he2023efficientdm,ViDiTQ2025,FP4DiT2025}. We fuse curvature and PCA into a normalized score, tier layers, assign W4 or W6 or W8 and keep full precision(FP) for the most sensitive.

\section{Related Work}
\label{sec:related work}

Unlike prior DiT PTQ that tunes precision or steps alone and relies on proxy latencies~\citet{bao2023one}, we couple both under one objective tied to measured hardware cost. We introduce drift guided pruning and a manifold-aware sensitivity score. 

\subsection{Post Training Quantization(PTQ) \& Timestep Optimization}
TRDQ improves quality and speed but adds rotations and per timestep knobs with unclear kernel costs~\citet{TRDQ2025}, while Qua2SeDiMo learns very low-weight precision only~\citep[p.~6155]{mills2025qua2sedimo}; surveys note gaps in hardware faithful deployment~\cite{Jayawant2025Computational}. We use integer kernels and jointly plan weights and activations, yielding clear accuracy–latency tradeoffs with a simple design.Few step restoration reduces sampling to about four steps but struggles at high resolution~\citep[p.~3452]{Han2023Enhancing}. We prune by measured teacher–student drift, keep a short protected tail, and select the top $k$ steps under a global budget to scale to harder settings. Fixed or uniform thinning often misses the error landscape and rarely co optimizes steps with precision~\cite{HTGDiT2025}. We co plan activation bits and the weight plan with step selection and evaluate using real integer kernels~\cite{FP4DiT2025}.


\subsection{Diffusion Transformer Architecture}
DiffI2I lightens the denoiser with a prior extractor and dynamic Transformer but needs two stage training~\citep[p.~2621]{Motetti2024Joint}. DMT~\citet[p.~3951]{Dong2024EQViT} transfers source information at one intermediate step for speed/quality, yet is sensitive to the chosen step. We propose PTQ without any retraining and make scheduling/precision choices data driven, which is drift and sensitivity aware, avoiding fragile step heuristics.

\subsection{U-Net Architecture}
DMFFT~\citet[p.~904]{Chai2025Diffusion} boosts U-Net~\cite{ronneberger2015u} quality by editing frequency components without extra training, trading flexibility for a fast plug in. Our method targets DiTs~\cite{bao2023one}and also works on UNet based diffusion models such as Stable Diffusion XL(SDXL)~\citet{podell2023sdxl}, and improvements come from selection (bits/steps) and hardware faithful kernels, irrespective of architecture.


\subsection{Layer Sensitivity and DAQ}
Single signal heuristics or fixed mixed precision rules misplace bits as statistics shift~\citep[p.~17537]{li2023q}~\cite{he2023ptqd,wu2024ptq4dit}. We fuse the magnitude signal $\sum x^2$ and PCA rank into one score, tier layers as robust to W4, mid to W6 or W8, fragile to FP, and seed a stable plan. Static tensor wise scales fail under timestep dependent outliers, prior work adopts dynamic or grouped scaling yet often lacks student aware calibration~\citep[p.~3]{chen2025}~\cite{vora2025ptq4adm,ViDiTQ2025,FP4DiT2025}. We add lightweight prehooks that compute per sample per timestep min max scales before grouped linears, tuned to the quantized student.

\section{DiffPro Overview}


\begin{figure}[!t]
  \centering
  \includegraphics[width=\columnwidth]{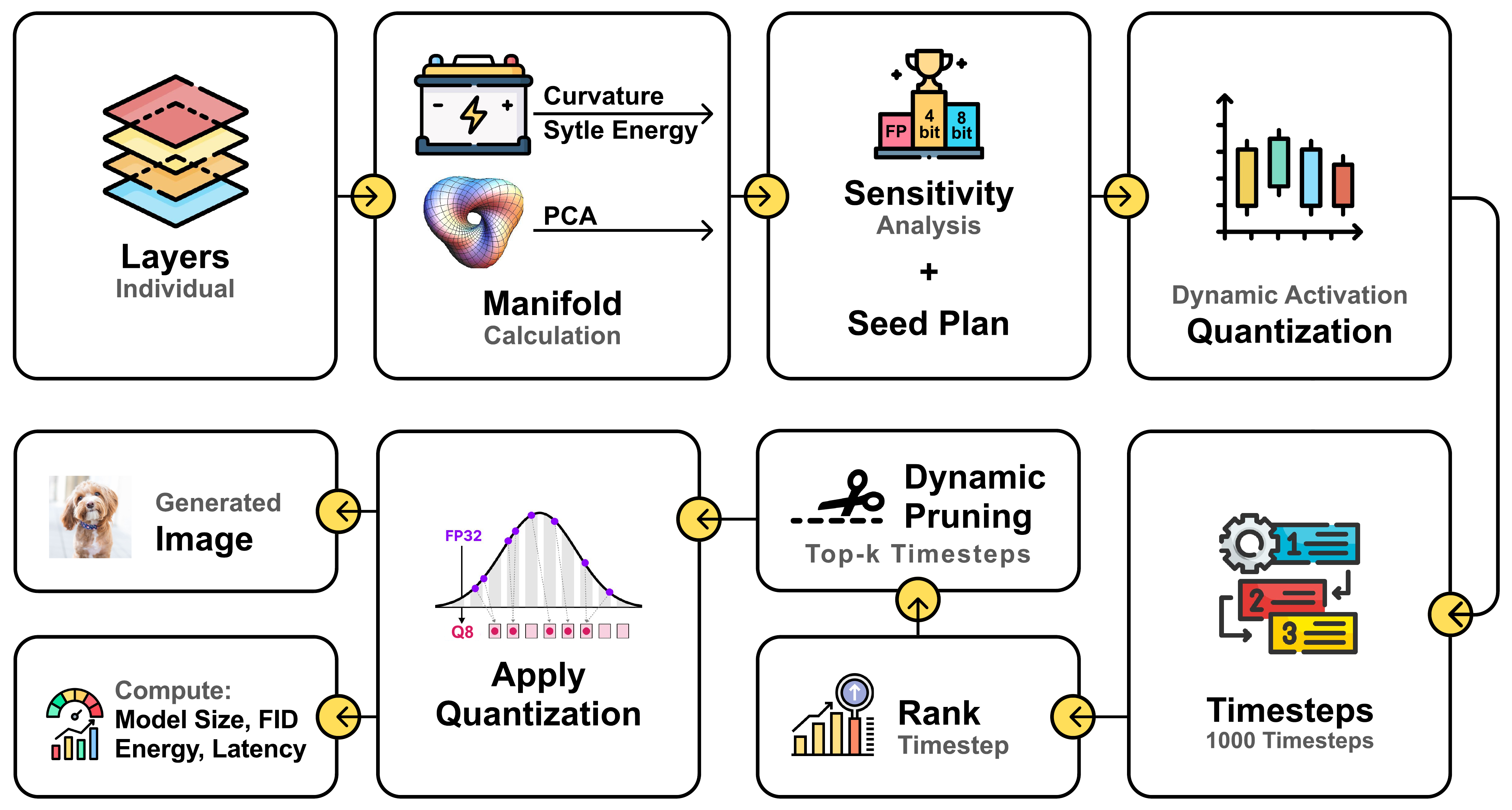}
  \vspace{-2em}
  \caption{DiffPro Overview:Joint precision optimization for efficient diffusion inference. The workflow analyzes layers, derives manifold and PCA signals, ranks sensitivity to seed a bit plan, applies 
  DAQ, prunes a 1000 step schedule with a protected tail, then quantizes and evaluates FID, latency, energy, and model size.}
  \label{fig:method}
  \vspace{-1em}
\end{figure}

\begin{algorithm}[]
    \vspace{0.5em}
  \caption{DiffPro (End to end,training free)}
  \label{alg:qdiffpro-core}
\begin{algorithmic}
  \STATE \textbf{Input:} Pretrained denoiser $f$; budgets $\mathcal{B}$; calibration set; group set $\mathcal{G}=\{32,64,128,192,288\}$
  \STATE \textbf{Calibration:} collect PCA stats (retain $95\%$); cache teacher features $(x_t,t,y)$
  \FOR{layer $\ell$}
    \STATE compute $S_\ell^{\mathrm{PCA}}$, drift with $\ell$ at low bits, phase variance, Jacobian energy, HF-bias
    \STATE $\mathcal{S}^{x}_\ell \gets$ blended composite; \quad $\mathcal{S}^{d}_\ell \gets$ D-PCA
    \STATE sweep small $(b,g)$ grid to get SRC slope $\mathcal{S}^{k}_\ell$ and knee $(b^\star_\ell,g^\star_\ell)$
    \STATE estimate noise $\mathcal{S}^{n}_\ell$ with $\sigma_\ell^2 \approx \Delta_\ell^2/12$ and local $\lVert \mathrm{Jac}_\ell \rVert_F^2$
    \STATE $S_\ell^\star \gets 0.4\,\mathcal{S}^{x}_\ell + 0.2\,\mathcal{S}^{d}_\ell + 0.25\,\mathcal{S}^{k}_\ell + 0.15\,\mathcal{S}^{n}_\ell$
  \ENDFOR
  \STATE tier layers by $S_\ell^\star$; freeze a fraction of high tier
  \STATE \textbf{Bit plan:} run Bit-ROI knapsack over W3/W4/W8 upgrades under $\mathcal{B}$ with guardrails $(b^\star_\ell,g^\star_\ell)$
  \STATE \textbf{Group sizes:} evaluate HCRG patterns per layer (screen B$=2$/T$=6$, confirm B$=6$/T$=12$); de-dup
  \STATE \textbf{DAQ:} enable Robust-DAQ (quantiles) with phase bins; add ORM
  \STATE \textbf{Pruning:} drift-aware step pruning with late protection
  \STATE \textbf{Post-calibration:} LNTC (bias $+$ LN); AdaRound-Lite on top-$k$ fragile layers
  \STATE \textbf{Screen $\rightarrow$ Confirm:} run cheap screen, then full confirm; print budget and score deltas vs.\ seed
  \STATE \textbf{Output:} final plan $\Pi^\star$
\end{algorithmic}
\end{algorithm}

 \noindent Our diffusion denoiser starts from pure noise and removes it over timesteps. Modern DiTs \citet{bao2023one} run in latent space via a VAE that maps pixels to a lower resolution four channel latent with a scaling constant \citet{pinheiro2021variational}. We use grouped weight quantization with per group scales and dynamic activation quantization. Smaller groups improve accuracy at modest overhead, while larger groups are cheaper but less accurate. Reverse diffusion can take hundreds of steps. Fewer steps are faster but riskier. We adopt a teacher–student setup with a full precision teacher and a compressed student. Unlike distillation training \citep{zhou2024simple}, the teacher audits a post training plan that jointly prunes steps and assigns precision with DAQ as shown in fig.~\ref{fig:method} and algorithm~\ref{alg:qdiffpro-core}. Unlike Q DiT \citet{chen2025}, \textsc{DiffPro} aligns dynamic activation with measured sensitivity, a drift budgeted bit and group plan, and a pruned schedule, while preserving $\mathrm{INT8}\times\mathrm{INT8}\rightarrow\mathrm{INT32}$ efficiency. The workflow spans 6 stages: 1.sensitivity, 2.per layer refinement, 3.dynamic activation learning, 4.pruning, 5.joint budgeted search \& 6.deployment.

 \noindent Algorithm~\ref{alg:qdiffpro-core} ranks layers by sensitivity and drift, runs small bit and group sweeps, estimates noise, and freezes the most fragile layers. It allocates bits to meet latency and memory targets, chooses group sizes, applies DAQ, prunes timesteps, performs light post calibration, confirms candidates, and outputs the final plan.

\section{DiffPro Details}
\subsection{Step~1: Calibration and Manifold Aware Sensitivity Estimation}
\label{step_1}

\noindent We run a brief calibration to gauge how fragile each linear layer is before quantization. Using a latent space DiT~\citet{bao2023one}, we VAE encode $256\times256$ images to $4\times32\times32$ latents~\citet{pinheiro2021variational}, add cosine scheduled noise at a few timesteps, and do a forward pass. Lightweight hooks on each linear layer record: (1) curvature energy $h_{\mathrm{diag}}=\sum x^2$ across batches, which flags layers more sensitive to rounding and (2) per group activation ranges (e.g., groups of 128 features), which provide data driven scales and a warm start for dynamic activation quantization. Then we run a per layer GPTQ~\cite{frantar2022gptq} preview guided by $h_{\mathrm{diag}}$ to pack \texttt{int8}/\texttt{int4} weights~\cite{frantar2022gptq}.


\noindent We add a PCA based manifold check. While the hooks run, we reservoir sample up to a few thousand input rows per layer to form $A\in\mathbb{R}^{N\times d}$. We run incremental PCA for up to $K$ components (e.g., $128$) and use the cumulative explained variance to read two values: $k_{95}$, the fewest components that reach $95\%$ variance, and $\mathrm{spill}=1-\mathrm{cum}(k_{95})$, leftover variance; if $95\%$ is not reached, we set $k_{95}{=}K$ and compute $\mathrm{spill}$. Small $k_{95}/d$ and low $\mathrm{spill}$ suggest robustness, large values signal quantization sensitivity. We combine these into a PCA sensitivity score in eq.~\ref{eq:s-pca}.

\begin{equation}
\label{eq:s-pca}
s_{\mathrm{PCA}} \;=\; 0.5\,\mathrm{spill} \;+\; 0.5\,\frac{k_{95}}{d}
\end{equation}

\noindent We then form one combined sensitivity score per layer. We normalize each layer’s mean $h_{\mathrm{diag}}$ to $[0,1]$ as $h_{\mathrm{diag}}^{\mathrm{norm}}$ and blend it with the PCA term in eq.~\ref{eq:s-combined}:
\begin{equation}
\label{eq:s-combined}
\mathrm{score} \;=\; \alpha \cdot s_{\mathrm{PCA}} \;+\; (1-\alpha)\cdot h_{\mathrm{diag}}^{\mathrm{norm}}, 
\qquad \alpha = 0.5
\end{equation}

We do not rely on a single signal such as curvature or block reconstruction error. Instead, our manifold-aware score \(s_M(l)\) is the explicit combination in eq.~\eqref{eq:s-combined} merging the curvature proxy \(h_{\mathrm{diag}}^{\mathrm{norm}}(l)\) with the PCA term \(s_{\mathrm{PCA}}(l)\) from eq.~\eqref{eq:s-pca}  all computed on the same latent space calibration batches and timesteps, yielding a concrete and reproducible layer ranking.We use  $\alpha$ = 0.5 to give equal importance to both score.

\begin{figure}[!t]
  \centering
  \includegraphics[width=0.8\columnwidth]{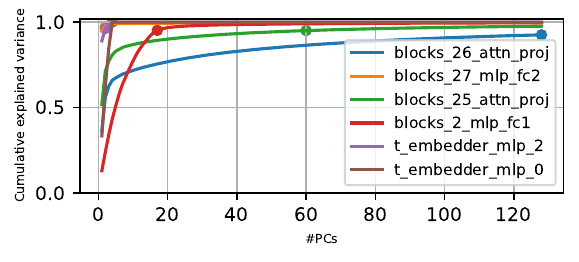}
  \vspace{-2em}
  \caption{PCA for the top  sensitive layers.}
  \label{fig:step1-1}
  \vspace{-1em}
\end{figure}

\noindent We rank layers by a PCA based score and assign three tiers for Steps~2--3. High score: safer settings like W8 or a few FP layers, smaller groups, often enable DAQ.Mid score: moderate settings W6 with default grouping.Low score: aggressive settings W4 with larger groups. Because we collect stats in the true latent space regime across early, mid, and late timesteps, these tiers match deployment, reduce the later search space, and preserve quality. PCA shows how concentrated activations are. If a few principal components(PCs) explain most variance, the layer is low rank and robust; if many are needed, it is sensitive. Fig~\ref{fig:step1-1} shows how much of a layer’s activation variance is captured as we add more PCs. The steep rise on the left means these layers are highly low rank. In practice, some layers reach about $95\%$ variance with $10$--$20$ components, while others need $30$--$40$. This guides tighter groups or lower precision for compressible layers, and slightly higher budgets for layers with slower rising curves.

\begin{figure}[!t]
  \centering
  \includegraphics[width=0.8\columnwidth]{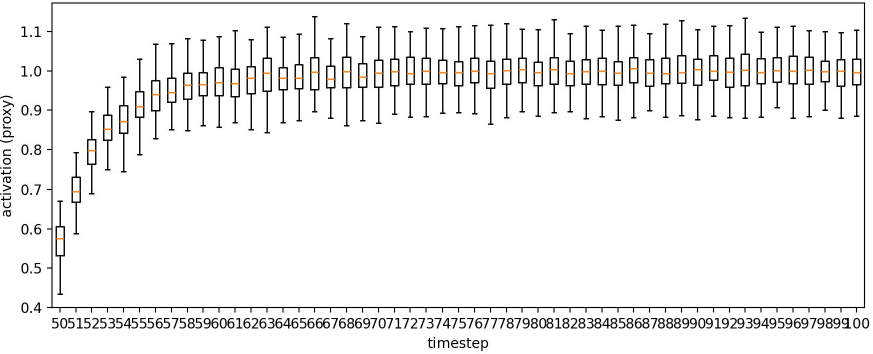}
  \vspace{-1.2em}
  \caption{Activation distribution over timesteps 50-100 for one chosen layer.}
  \label{fig:step1-2}
  \vspace{-1em}
\end{figure}

\noindent Fig.~\ref{fig:step1-2} plots an activation proxy for one layer as the diffusion timestep increases from $50$ to $100$. Early steps show rapid structure formation; after $t\approx70\text{--}80$ the behavior stabilizes. It shows that for calibration, including early–mid steps is required where distributions shift as well as later ones i.e. where it stabilizes.

\begin{figure}[!t]
  \centering
  \includegraphics[width=0.8\columnwidth]{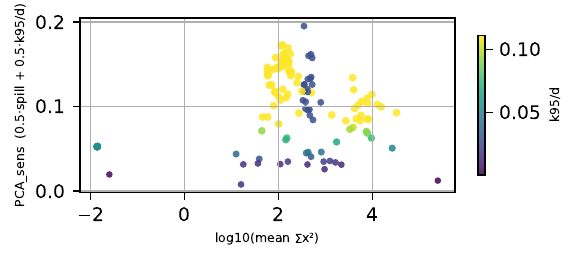}
  \vspace{-1.2em}
  \caption{Scatter:$\log_{10}\!\bigl(\mathrm{mean}\,\sum x^{2}\bigr)$ vs. \ $\mathrm{PCA\_sensitivity}$ (size/color $\approx k_{95}/d$).}
  \label{fig:step1-4}
  \vspace{-1em}
\end{figure}

\noindent Fig.~\ref{fig:step1-4} plots one dot per layer. $x$ is $\mathrm{mean}\,\sum x^{2}$ (more right $\rightarrow$ larger magnitude) and $y$ is $\mathrm{PCA\_sensitivity}$ (higher $\rightarrow$ more complex).Dot size/color encode $k_{95}/d$ (bigger/yellower $\rightarrow$ needs more principal components).
Upper right layers are hardest; lower left are easiest.
High $y$ but moderate $x$ signals geometry risk; high $x$ but low $y$ signals magnitude risk.


\begin{figure}[!t]
  \centering
  \includegraphics[width=0.8\columnwidth]{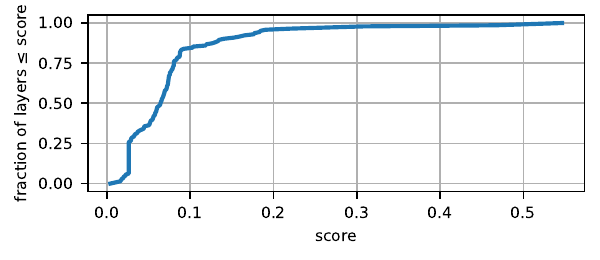}
  \vspace{-1.5em}
  \caption{CDF of the combined score.}
  \label{fig:step1-5}
\end{figure}

Fig.~\ref{fig:step1-5} shows how the final cumulative distribution function(CDF) scores are distributed across all layers. For any score on the $x$ axis, the $y$ value tells  what fraction of layers are at or below this score. 

\begin{figure}[!t]
  \centering
  \includegraphics[width=0.8\columnwidth]{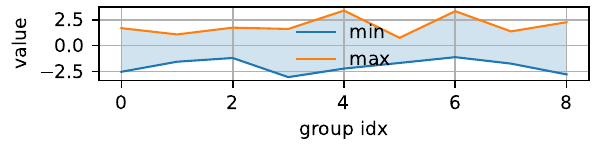}
  \vspace{-1.2em}
  \caption{Activation envelope per 9 groups (min/max) in a block.}
  \label{fig:step1-6}
\end{figure}

Fig.~\ref{fig:step1-6} plots the per group minima and maxima of the inputs for one layer, the shaded band is their envelope. A wide envelope across many groups means activations vary a lot and we need more buckets or DAQ. A narrow envelope means the activations are easy to quantize.

\begin{figure}[!t]
  \centering
 \includegraphics[width=0.8\columnwidth]{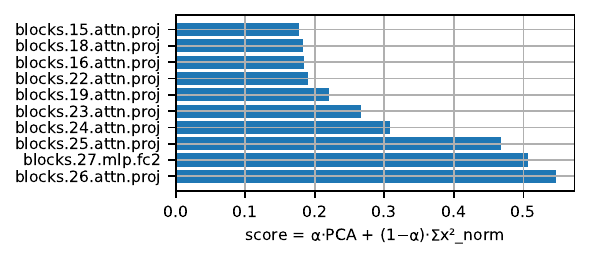}
 \vspace{-1.5em}
  \caption{ Top-10 layers by combined score.}
  \label{fig:step1-7}
\end{figure}

 Finally, fig.~\ref{fig:step1-7} shows the top 10 layers with the highest overall risk, sorted from most to least.

 The figures illustrated in step~1 give a blueprint for quantizing per layer weights and activations.

\subsection{Step~2: Per Layer Bit and Course Group Size Refinement}
\label{step_2}


\noindent From Step~1, we convert layer sensitivity stats into a tested quantization plan by combining PCA manifold sensitivity and Hessian diagonal input sensitivity. Using the combined score, we set a seed bit plan: low score layers use W4 with coarse groups 192--288, mid score layers use W8 with default groups, and high score layers use FP16 with small groups 32--64. This reflects nonuniform layer fragility and assigns precision across depth accordingly.


\noindent We refine post training with an evolutionary search along with GPTQ~\citet{frantar2022gptq}. We load the full precision teacher once, run it on a small calibration set of latent noise , timesteps , and  class labels from Step~1, and cache outputs to avoid repeats. Each generation mutates bit widths and group sizes around the seed while freezing the most sensitive layers. For each candidate, we build a student with GPTQ~\citet{frantar2022gptq}: approximate the per layer Hessian $H \approx X^\top X$ from a few batches, solve with a Cholesky column~\citet{smith1995differentiation} method, and quantize weights row wise and group wise to integer tensors $qW$ with per group scale and zero point. We support Linear and convolution layers by unfolding convolution inputs before computing $H$, then swap in wrappers that store integer weights and dequantize on the fly during scoring.

\noindent In eq.~\ref{eq:step2-drift} we score each student by the mean squared difference between student and teacher noise predictions:
\begin{equation}
\label{eq:step2-drift}
\mathcal{L}_{\text{drift}} \;=\; \mathbb{E}_{(x_{t}, t, y)} \left\| \epsilon_{\text{student}}(x_{t}, t, y) \;-\; \epsilon_{\text{teacher}}(x_{t}, t, y) \right\|_{2}^{2}
\end{equation}
To reduce cost, we use successive halving: we first test candidates with a few batches/timesteps, then re-test the best ones with stronger settings. We skip duplicate plans via hashing. Across generations, we keep the elites with the lowest drift and mutate them until convergence, which yields a refined bit plan. We then move precision to where it helps most, fragile blocks get more bits or smaller groups, while robust blocks stay compact.

\begin{figure}[t]
  \centering
  \includegraphics[width=0.8\columnwidth]{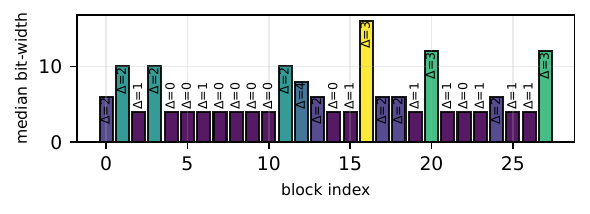}
  \vspace{-1.5em}
  \caption{Best Bit-Width by Block (Median, $\Delta=$Changed Layers.)}
  \label{fig:step2-1}
  \vspace{-1em}
\end{figure}

\noindent Fig.~\ref{fig:step2-1} summarizes final precision across depth. Each bar is a transformer block, its height is the median bit width of the quantized layers in that block, and color intensity encodes the same value, with lighter meaning higher precision. The symbol $\Delta$ above a bar counts layers whose bit assignment changed from the Step~1 seed. Middle blocks with dark purple and $\Delta\approx0$ stayed stable, confirming robustness at low precision.

\begin{figure}[t]
  \centering
  \includegraphics[width=0.8\columnwidth]{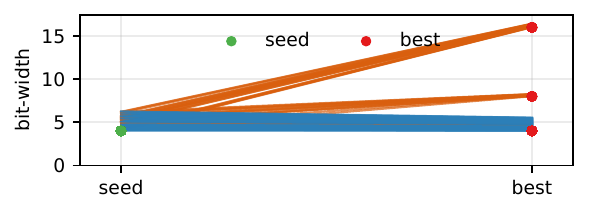}
  \vspace{-1.5em}
  \caption{ Bit-Width Changes per Layer, from seed to best plan (Slope Chart).}
  \label{fig:step2-4}
  \vspace{-1em}
\end{figure}

\noindent Fig.~\ref{fig:step2-4} shows bit width changes per layer from the seed to the optimized plan. Green dots at $x=\text{seed}$ and red dots at $x=\text{best}$ mark each layer. Horizontal blue segments mean layers stayed at $4$ bit. Orange rising lines mark upgrades such as $4\rightarrow8$ or $4\rightarrow16$. The thick band of blue near the bottom shows most layers remained efficiently quantized.

\begin{figure}[t]
  \centering
  \includegraphics[width=0.8\columnwidth]{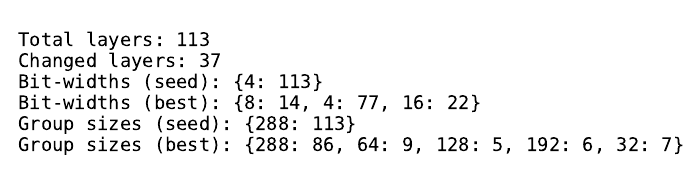}
  \vspace{-1.2em}
  \caption{Summary of layers, bits, and groups refinement.}
  \label{fig:step2-5}
  \vspace{-1em}
\end{figure}

\noindent Fig.~\ref{fig:step2-5} shows the final model after refinement. The network has 113 layers, and 37 changed from the seed plan. The seed used W4 and a group size of 288 for all layers. The best plan shifts  77 layers at W4, 14 at W8, and 22 at FP16, with group sizes spread across 86 at 288, 9 at 64, 5 at 128, 6 at 192, and 7 at 32. Most layers stay compact, while a targeted subset gains higher precision and smaller groups to recover quality, yielding a balanced accuracy–efficiency tradeoff.

 In Step~2, the insights got from Step~1 are utilized to come up with the bit width refinement.

\subsection{Step~3: Dynamic Activation Quantization (DAQ)}
\label{step_3}

We add DAQ in Step~3 to keep low bit inference stable as activations change over timesteps. For layer $l$ at step $t$, let $a^{(l)}_t\!\in\!\mathbb{R}^{B\times C\times\cdots}$ be per-sample activations. We split channels into groups of size $g$ and flatten each group to $v_{btk}$ for sample $b$ and group $k$. We compute a robust symmetric INT8 scale per \emph{sample}, \emph{timestep}, and \emph{group} in eq.~\ref{eq:daq}:
\begin{equation}
\begin{aligned}
\tau_{btk} &\coloneqq \operatorname{Percentile}\!\big(|v_{btk}|,p\big),\\
\alpha_{btk} &\coloneqq \frac{\tau_{btk}}{2^{\,b-1}-1},\quad b=8,\\
\hat v_{btk} &\coloneqq \alpha_{btk}\,\mathrm{round}\!\left(
  \frac{\mathrm{clip}\!\left(v_{btk},-\tau_{btk},\tau_{btk}\right)}{\alpha_{btk}}
\right)
\end{aligned}
\label{eq:daq}
\end{equation}

with $p{=}99.9$ for outlier clipping and zero-point $0$. Lightweight pre-hooks compute these scales right before each grouped linear; weights are GPTQ~\cite{frantar2022gptq} packed, and grouped linears run as $\mathrm{INT8}\times\mathrm{INT8}\rightarrow\mathrm{INT32}$ GEMMs. DAQ is applied to the inputs of $Q$, $K$, $V$, and the attention output projection, and to both multi layer perceptron(MLP) layers, LayerNorms and residual adds remain in higher precision. This design adapts to changing statistics while avoiding per timestep tables and keeping kernels hardware friendly.

Prior DiT~\cite{bao2023one} PTQ often uses fixed or layer only activation scales, proxy latencies, and lacks student aware calibration. Our DAQ is tuned to the quantized student, and it is integrated into a joint co optimization of timesteps and per layer precision under a hardware measured latency budget, guided by drift aware scheduling and manifold aware sensitivity from Steps~1--2. We run three quick checks: DAQ on/off latency, where there is a small overhead with fused kernels, teacher--student drift using shared synthetic latents, gaining low but non zero mean squared error(MSE), and a small sample Fréchet Inception Distance(FID)~\cite{yu2021frechet} test before full evaluations.

\begin{figure}[t]
  \centering
  \includegraphics[width=0.8\columnwidth]{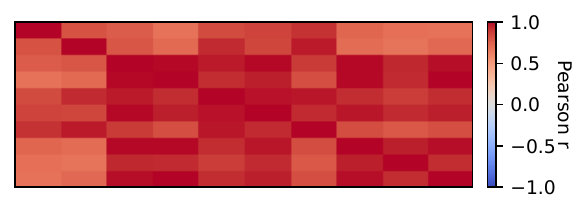}
  \vspace{-1.5em}
  \caption{Layer-wise correlation heatmap of $\mathrm{std}(t)$ for DiT-XL/2,256X256.}
  \label{fig:step3-1}
  \vspace{-1em}
\end{figure}

\noindent Fig.~\ref{fig:step3-1} maps how layer activation variability aligns over timesteps. Each cell gives the Pearson correlation between the timewise standard deviation of two layers. The deep red field shows that most layers rise and fall together under the same noise schedule in DiT~\cite{bao2023one}. Lighter vertical or horizontal bands mark groups that deviate. The bright diagonal is self correlation, and the pale bands flag layers with distinct temporal behavior that counts for special DAQ handling or freezing.

\begin{figure}[t]
  \centering
  \includegraphics[width=0.8\columnwidth]{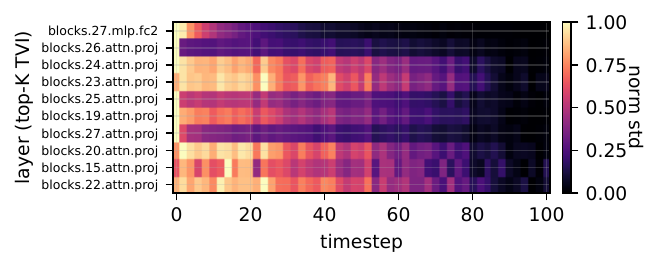}
  \vspace{-1.5em}
  \caption{Row normalized std heatmap for top-$K$ layers by temporal variability.}
  \label{fig:step3-2}
  \vspace{-1em}
\end{figure}

\noindent In fig.~\ref{fig:step3-2}, each row represents a layer chosen by a temporal variability index, each column is a timestep, and colors show that layer’s row normalized standard deviation(std). Bright early columns fading to darker late columns reveal a common pattern that activations are more volatile early and stabilize as denoising progresses. 

\begin{figure}[t]
  \centering
  \includegraphics[width=0.8\columnwidth]{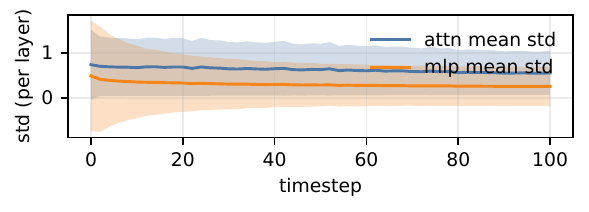}
  \vspace{-1.5em}
  \caption{ Attention vs.\ MLP ribbons over time.}
  \label{fig:step3-3}
  \vspace{-1em}
\end{figure}

\noindent Fig.~\ref{fig:step3-3} shows the mean standard deviation of activations across attention and MLP layers at each timestep, with ribbons for plus or minus one std across layers. Attention starts higher and remains above MLP, and both bands decline over time as denoising moves from noisy to clean states.

\begin{figure}[t]
  \centering
  \includegraphics[width=0.8\columnwidth]{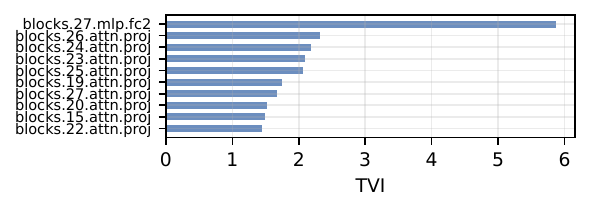}
  \vspace{-1.5em}
  \caption{Top-20 layers by Temporal Variability Index (TVI).}
  \label{fig:step3-4}
  \vspace{-1em}
\end{figure}

\noindent Fig.~\ref{fig:step3-4} ranks layers using an index that combines average std and its dispersion over time 
The late block attention projections and the final MLP head are the most time sensitive components and prime targets for DAQ.

These show that variability is largest early in the trajectory and usually higher in attention blocks than in MLPs.DAQ provides the  temporal adaptivity for activation quantization.

\subsection{Step~4: Budgeted Timestep Selection (Pruning)}
\noindent After the bit width refinement, we pick which timesteps to keep using a small data driven score. We keep a fixed full precision teacher $f_{\theta}$ and a quantized student $f_{\hat{\theta}}$ (from Steps~2--3). In eq.~\ref{eq:delta-t}, for each candidate step $t \in \mathcal{C} \subseteq \{0,\ldots,T-1\}$, we take tiny mini-batches $(x_t,y)$ in latent space and measure how far the student is from the teacher:
\begin{equation}
\label{eq:delta-t}
\delta(t) \;=\; \mathbb{E}_{(x_t,y)} \bigl\| f_{\hat{\theta}}(x_t,t,y) \;-\; f_{\theta}(x_t,t,y) \bigr\|_{2}^{2}
\end{equation}
This gives one number per timestep. It is usually larger near the start and smaller near the end. We always keep the last $\rho$ part of the timeline, typically $\rho=0.2$ to keep late steps stable. From the remaining steps, we sort by $\delta(t)$ big to small and take the Top-$K$ until we fill the budget. Formally, with
\[
\mathcal{T}_{\mathrm{tail}} \;=\; \{\, t \in \mathcal{C} : t/T \ge 1-\rho \,\},
\]
and letting $\operatorname{TopK}(\cdot,\delta(\cdot),m)$ pick the $m$ largest by $\delta$, as eq. ~\ref{eq:schedule} shows our final schedule is
\begin{equation}
\label{eq:schedule}
\mathcal{S} \;=\; \mathcal{T}_{\mathrm{tail}} \;\cup\; \operatorname{TopK}\!\bigl(\,\mathcal{C}\!\setminus\!\mathcal{T}_{\mathrm{tail}},\, \delta(\cdot),\, k - |\mathcal{T}_{\mathrm{tail}}| \bigr),
\qquad |\mathcal{S}| = k
\end{equation}

\begin{figure}[!t]
  \centering
  \includegraphics[width=0.8\columnwidth]{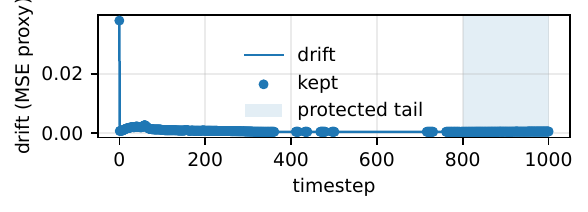}
  \vspace{-1.5em}
  \caption{ Kept Timesteps According to drift:MSE Vs Timestep}
  \label{fig:step4-1}
  \vspace{-1em}
\end{figure}

\begin{figure}[!t]
  \centering
  \includegraphics[width=0.8\columnwidth]{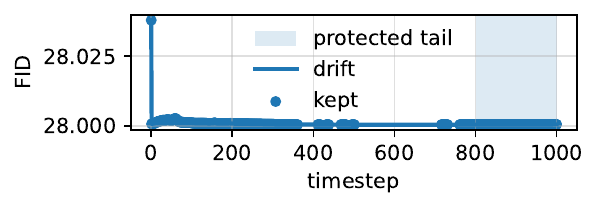}
  \vspace{-1.5em}
  \caption{ Kept timesteps according to drift:FID Vs Timestep}
  \label{fig:step4-12}
  \vspace{-1em}
\end{figure}

We define teacher--student drift per diffusion step $t$ as the batch mean $\ell_2$ error on noise predictions using the same latents, i.e., $D_e(t)=\tfrac{1}{B}\sum_{i=1}^{B}\!\left\|\hat{\epsilon}_{\text{stu}}(z_i,t)-\hat{\epsilon}_{\text{teach}}(z_i,t)\right\|_2^2$. Steps are ranked by $D_e(t)$ where higher means more fragile, we keep high drift steps, prune low drift ones, and always retain a protected tail of late steps. In our plots, fig.~\ref{fig:step4-1} and ~\ref{fig:step4-12} drift spikes near $t\!\approx\!0$, exhibits a small early hump, then decays moderate around $t\!\approx\!20$--$300$ and very low thereafter with dots marking kept steps and a shaded band indicating the always kept tail.

\noindent We report Lorenz and Gini plots to show how much error mass the kept steps cover and where they occur in time. In our runs, the kept set covers about two thirds at $k=500$. The method is practical and model agnostic, it measures the deployed student with bits, groups and DAQ applied, uses a small latent calibration stream and scales linearly with $|\mathcal{C}|$ times the mini batch count per step. A protected tail secures late stage fidelity, while top $k$ targets early steps where errors concentrate. We then plug $\mathcal{S}$ into the sampler to trade steps for latency with a clear, data backed guarantee of retained teacher student error.


\begin{figure}[!htpb]
  \centering
  \includegraphics[width=0.8\columnwidth]{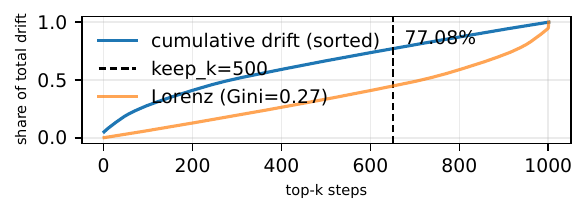}
  \vspace{-1.5em}
  \caption{Sorted drift elbow at kept 500 timesteps.}
  \label{fig:step4-2}
  \vspace{-1em}
\end{figure}

\noindent Fig.~\ref{fig:step4-2} shows that timesteps are sorted by drift and FID, and the curve shows the cumulative share of total drift versus the Top-$K$ steps. The dashed line at $\mathrm{keep\_k}=500$ lands around $77.68\%$, meaning the 500 kept steps capture about two thirds of all drift. 

\vspace{1em}
\subsection{Step~5: Joint, Budget-Aware Plan Search}
\begin{algorithm}[!t]
\caption{Joint Planning with bits \& timesteps and Deployment with DAQ}
\label{alg:joint+deploy-compact}
\begin{algorithmic}[1]
\STATE \textbf{Inputs:} steps $\mathcal{T}$ with drift $D_e(t)$ and per step latency $c_{\text{step}}(t)$; layers $\mathcal{L}$ with bit options $\mathcal{B}_\ell$, costs $c_{\text{lat}}(\ell,b),c_{\text{mem}}(\ell,b)$, sensitivity $s_M(\ell)$; budgets $B_{\text{lat}},B_{\text{mem}}$; protected-tail fraction $\rho$; DAQ policy $\mathcal{P}_{\mathrm{DAQ}}$; VAE; checkpoint $\mathcal{C}$.
\STATE \textbf{Outputs:} bitplan $\Pi^\star:\mathcal{L}\!\to\!\mathcal{B}_\ell$, kept steps $\mathcal{K}^\star\!\subseteq\!\mathcal{T}$, images.
\vspace{0.2em}
\STATE \textbf{Phase A: Budgeted joint planning.}
\STATE Tail rule: $\textsf{Tail}\!\leftarrow$ last $\lfloor\rho|\mathcal{T}|\rfloor$ indices; always enforce $\textsf{Tail}\subseteq\mathcal{K}$.
\STATE Init: $\Pi(\ell)\!\leftarrow\!\max\mathcal{B}_\ell$, $\mathcal{K}\!\leftarrow\!\mathcal{T}$; compute $C_{\text{lat}},C_{\text{mem}}$.
\STATE Sort prunable steps $\mathcal{R}\!=\!\mathcal{T}\setminus\textsf{Tail}$ by $D_e(t)$ asc; build max-heap $H_{\text{bit}}$ of moves $(\ell{:}b\!\to\!b')$ scored $S_{\text{bit}}=\Delta c_{\text{lat}}/(\varepsilon{+}s_M(\ell))$.
\WHILE{$C_{\text{lat}}>B_{\text{lat}}$ \OR $C_{\text{mem}}>B_{\text{mem}}$}
  \STATE Step candidate: $t^\star\!=\!\text{first}(\mathcal{R})$, $S_{\text{step}}=c_{\text{step}}(t^\star)/(\varepsilon{+}D_e(t^\star))$.
  \STATE Bit candidate: $(\ell^\star{:}b\!\to\!b')\!=\!\text{top}(H_{\text{bit}})$ s.t. $C_{\text{mem}}{-}\Delta c_{\text{mem}}\!\le\!B_{\text{mem}}$.
  \IF{$S_{\text{step}}>S_{\text{bit}}$} \STATE $\mathcal{K}\!\leftarrow\!\mathcal{K}\setminus\{t^\star\}$; $\mathcal{R}\!\leftarrow\!\mathcal{R}\setminus\{t^\star\}$; $C_{\text{lat}}{-}\!=c_{\text{step}}(t^\star)$
  \ELSE \STATE $\Pi(\ell^\star)\!\leftarrow\!b'$; update $C_{\text{lat}},C_{\text{mem}}$; push next cheaper move for $\ell^\star$ if any
  \ENDIF
\ENDWHILE
\STATE Set $\Pi^\star\!\leftarrow\!\Pi$, $\mathcal{K}^\star\!\leftarrow\!\mathcal{K}$.
\vspace{0.2em}
\STATE \textbf{Phase B: Deployment with DAQ and pruned schedule.}
\STATE Load $\mathcal{C}$; wrap layers per $\Pi^\star$; init sampler on $\mathcal{K}^\star$.
\FOR{$t \in \mathrm{reverse}(\mathcal{K}^\star)$}
  \STATE Bin $b\!\in\!\{\text{early, mid, late}\}$; set $a_{\mathrm{bits}}\!\leftarrow\!\mathcal{P}_{\mathrm{DAQ}}(b)$; forward with quantized weights and activations; update sampler.
\ENDFOR
\STATE Decode with VAE; save outputs.
\vspace{-0.1em}
\STATE \textbf{Complexity:} Phase A $=O(T\log T + L\log L)$ (sorts+heaps); Phase B linear in $|\mathcal{K}^\star|$.
\end{algorithmic}
\end{algorithm}
\noindent Step~5 selects one deployable plan that meets a fixed budget. The plan combines a weight quantization plan from Step~2, a dynamic activation policy over time from Step~3, and a pruned sampling schedule from Step~4, as shown in algorithm~\ref{alg:joint+deploy-compact}. We keep the teacher DiT \citet{bao2023one} in full precision and build a student that runs real integer math. Every linear layer uses an $\mathrm{INT8}\times\mathrm{INT8}\rightarrow\mathrm{INT32}$ kernel with per outputchannel weight scales, and activations are quantized on the fly per sample and per timestep under the DAQ policy. Different parts of the diffusion trajectory use different precision through three time bins: early, mid, and late.

\noindent The search starts from the best bit plan from Step~2, the DAQ policy from Step~3, and the kept timesteps from Step~4. It varies DAQ bits for selected layers and slightly shortens or lengthens the kept-timestep list, while keeping weights fixed at $\mathrm{INT8}$ from GPTQ \citet{frantar2022gptq} to ensure true deployable kernels.

\noindent Step~5 performs a joint budgeted search over these deployable plans. Each candidate is evaluated on a small calibration set with a score that adds teacher–student error to penalties for latency and compute under fixed limits on milliseconds and BitOps. An evolutionary loop keeps the best candidates, mutates DAQ bits and the kept schedule, and outputs the final plan measured with real integer kernels explained in phase A of algorithm~\ref{alg:joint+deploy-compact}

\noindent We compute the score on cached teacher features and student runs. We cache the teacher outputs once, time the student with CUDA, estimate compute with a BitOps proxy, and combine the score as eq. ~\ref{eq:step5-score}

\begin{equation}
\label{eq:step5-score}
\mathrm{score}
= \mathrm{MSE}
+ \lambda\,\text{latency\_penalty}
+ \mu\,\text{bitops\_penalty}.
\end{equation}
Here $\lambda$,$\mu$= 0.5 ,to put more emphais on latency or Bitops or 1. The evolutionary loop keeps the top candidates each round, mutates them across rounds and outputs the best plan.

\begin{figure}[!t]
  \centering
  \includegraphics[width=0.8\columnwidth]{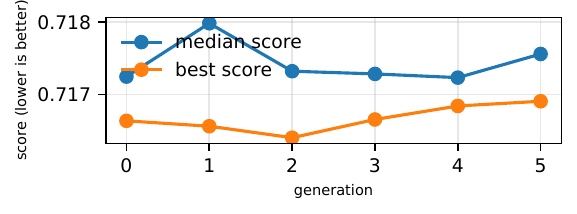}
  \vspace{-1.5em}
  \caption{Evolution: score by generation.}
  \label{fig:step5-1}
  \vspace{-1em}
\end{figure}

\noindent Fig~\ref{fig:step5-1} plots the median candidate per generation and the best candidate. The best score drops to about 1.011 in the first generation, then plateaus with minor fluctuations and a slight recovery by generations four to five, indicating early improvement with only small later gains.

\begin{figure}[t]
  \centering
  \includegraphics[width=0.6\columnwidth]{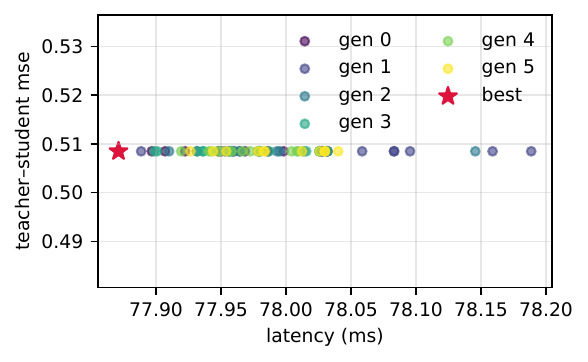}
  \vspace{-1.5em}
  \caption{Mean squared error(MSE) vs.\ latency by generation.}
  \label{fig:step5-2}
  \vspace{-2em}
\end{figure}

\noindent In, fig \ref{fig:step5-2}  most points cluster near $\sim 111$--$112$ ms latency with teacher--student MSE $\approx 0.509$. 
The red star (best by score) sits on the low-latency edge $\sim 110$--$111$ ms with slightly better MSE than its neighbors, so the search chose a small quality win without extra latency. 

\subsection{Step 6: Deployment and Sampling}

\noindent In this stage, at each step, the activation bits are set for each layer according to which bin $t$ falls in early/mid/late. At inference time, the architecture is instantiated and wrapped with quantized linear layers according to the final bit plan. A lightweight DAQ runtime sets activation bit width before each forward pass based on the current timestep’s bin. A small function is attached that sets each layer’s activation bits from the predefined policy for the bin containing $t$. The sampler iterates the pruned list $\mathcal{K}$ in reverse order, applying one reverse diffusion update per step. Phase B in algorithm~\ref{alg:joint+deploy-compact}, explains this in details.


\section{Implementation}
\noindent We use PyTorch implementations of a class conditioned DiT-XL/2~\cite{bao2023one} and an image conditioned U-Net from SDXL~\cite{podell2023sdxl}. Experiments run at 256×256 on ImageNet~\cite{deng2009imagenet} and CelebA-HQ~\cite{karras2018celebAHQ} and at 32×32 on CIFAR-10 and CIFAR-100~\cite{krizhevsky_cifar_2009}. We sample 500 images per model with DDIM~\cite{song2020denoising}. DiT~\cite{bao2023one} uses 1{,}000 steps, SDXL uses 100 steps. We use DDIM with 1000 steps on the original 1000 step grid to exactly match the teacher’s DDPM~\cite{ho2020denoising} noise schedule. For class conditioned DiT~\cite{bao2023one},around 500 timesteps were kept, and image conditioned 50 timesteps. Classifier-free guidance scale is 1.0 for DiT-XL/2 and 3.0 for SDXL, and we report FID~\cite{yu2021frechet} on 500 samples. Tests ran on NVIDIA A30 24 GB GPUs with CUDA 12.1.1 and PyTorch 2.1.2 via the cuBLASLt and CUTLASS INT8 path. All DiT linear and projection layers use \texttt{INT8×INT8→INT32} GEMM. We pack weights with GPTQ and apply per sample and per timestep DAQ using a 512 example calibration set from Step~1. Latency is end to end over the full denoising loop, including DAQ scale computation, requantization and dequantization, residual additions, and layer normalization. We use synchronized CUDA events, average 100 runs after warm up, and report batch sizes 1 and 4 in channels last format. TensorRT is not used, and results reflect the PyTorch INT8 path.

\section{Evaluation}

\subsection{Results on Different Datasets and Comparison with Baselines}

DiffPro is tested on different datasets and  the recorded  throughput, model size, and energy are depicted in table~\ref{tab2}.

\begin{table}[!htpb]
\caption{Results on different datasets. 
}
\label{tab2}
\vskip 0.05in
\begin{center}
\begin{small}
\begin{sc}
\vspace{-1em}
\resizebox{\columnwidth}{!}{%
\begin{tabular}{lccccc}
\toprule
Dataset & FID $\downarrow$ & Throughput(Sec/Image) & Reduced Size(MB)$\downarrow$ & FP Size(MB) & Energy(J) \\
\midrule
Cifar-10  & 20.02  & 0.21 & 397.24  & 2575.42  & 298.95 \\
CelebA    & 10.95  & 0.47 & 397.24  & 2575.42  & 410.26 \\
Cifar-100  & 12.39  & 0.39 & 397.24  & 2575.42  & 360.78 \\
Imagenet  & 28.05  & 0.42 & 397.24  & 2575.42 & 360.05 \\
\bottomrule
\end{tabular}%
}
\end{sc}
\end{small}
\end{center}
\vskip -0.05in
\vspace{-1em}
\end{table}

We compare the proposed DiffPro method with existing PTQ methods in table~\ref{tab3}. PTQ4DM ~\citep[p.~4]{vora2025ptq4adm},Q-Diffusion~\citep[p.~17544]{li2023q} and PTQD~\cite{he2023ptqd} are three PTQ methods designed for U-Net based diffusion models. Q-DiT~\citep[p.~28314]{chen2025} is specially tailored for the Diffusion Transformer architecture. DiﬀuseVAE~\citet{pandey2022diffusevae} conducts a study on optimization of timesteps.

\begin{table}[!htpb]
\caption{Comparison with baselines.}
\label{tab3}
\vskip 0.25in
\begin{center}
\begin{small}
\begin{sc}
\vspace{-1em}
\resizebox{\columnwidth}{!}{%
\begin{tabular}{lccccc}
\toprule
Name & FID$\downarrow$ & FP Size(MB) & Reduced Size(MB) & Size Reduction(\%) \\
\midrule
PTQ4DM  & 89.78  & \_\_ &  \_\_ & 68\%   \\
Q-Diffusion & 54.95 & 1096.2  & 274.1 & 75\%   \\
PTQD   & 26.64  & 1403.35 & 334.51 &   76\%  \\
DiﬀuseVAE  & 23.20  & \_\_ & \_\_  & \_\_ \\
QDiT  & 21.59  & \_\_ & \_\_  & \_\_ \\
\textbf{DiffPro(DiT)} & \textbf{28.05}   & \textbf{2575.42} & \textbf{397.24} & \textbf{84\%} \\
\textbf{DiffPro(UNet)} & \textbf{29.54}  & \textbf{12877.10} & \textbf{2317.87} & \textbf{82\%} \\
\bottomrule
\end{tabular}%
}
\end{sc}
\end{small}
\end{center}
\vskip -0.05in
\vspace{-1em}
\end{table}

\noindent We tested PTQ on DiT-XL/2 at $256{\times}256$. With pruned steps, DiffPro achieves a $6.25{\times}$ reduction in model size and a 50\% reduction in timesteps with competitive FID. Although~\citet{he2023ptqd},~\citet{pandey2022diffusevae}, and~\citet{chen2025} report slightly lower FID~\cite{yu2021frechet} due to evaluation on 5k images, DiffPro delivers the largest reduction in model size, and~\citet{chen2025} and~\citet{pandey2022diffusevae} do not report size reduction. DiffuseVAE~\citet{pandey2022diffusevae} attains a slightly better FID with one-thousand timesteps, while \textsc{DiffPro} reaches comparable quality with half as many timesteps. We also validate on U-Net based SDXL~\cite{podell2023sdxl} using class labels as prompt and observe similar trends in FID and compression. For U-Nets we insert DAQ before convolutions and attention projections, rank blocks by stage, handle cross and temporal attention like DiT query, key, value, and output, and freeze only a few of the most sensitive layers, with the main differences in the layer set and group sizes. We measure energy per image by integrating GPU power from NVML over the denoising loop and dividing total Joules by the number of images. We measure latency with synchronized CUDA using a similar GPU and batch settings.

Sufficient experiments demonstrate that the proposed DiffPro method can quantize large scale Diffusion Models (2-12 GB) with different architectures without noticeable performance degradation.

\begin{figure}[!t]
  \centering
  \includegraphics[width=0.8\columnwidth]{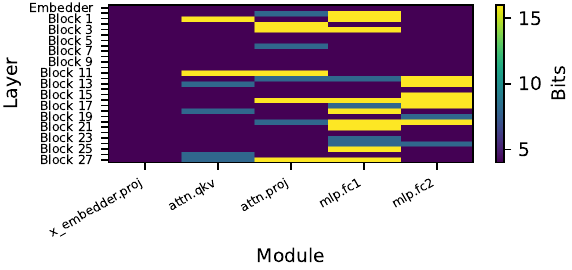}
  \vspace{-1.5em}
  \caption{Bit per layer heatmap according to best seed plan.}
  \label{fig:step_ab2}
  \vspace{-1em}
\end{figure}

\begin{figure}[!t]
  \centering
  \includegraphics[width=0.8\columnwidth]{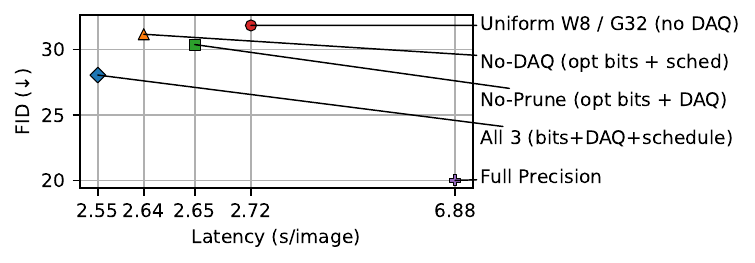}
  \vspace{-0.5em}
  \caption{Ablation of DiffPro components: Impact of bitplan, DAQ, and pruned schedule on FID and latency.}
  \label{fig:step_ab3}
  \vspace{-1em}
\end{figure}

Fig.~\ref{fig:step_ab2} shows the final bits per layer best plan heatmap showing assigned precisions (W4/W8/W16) across  DiT blocks/modules, with color indicating bit width. And fig.~\ref{fig:step_ab3} shows quality–latency tradeoff: DiffPro baseline with all 3 (bits+DAQ+schedule) achieves the lowest latency at similar FID, while full precision is much slower.

\subsection{Ablation Study}

\noindent To validate the effectiveness of our components, we conduct an in depth ablation study in which we evaluate quality with FID score\cite{yu2021frechet}, which is lower is better, and tested speed as average per image latency on Imagenet~\cite{deng2009imagenet}. We generate 500 images to compute FID on a cleaned reference set, and we keep calibration batches fixed across all variants for consistency. As the baseline, we use the full Step 5 pipeline and then gradually remove specific parts to isolate their contributions. Concretely, we study three variants: 
\noindent(i) {No-DAQ}, which fixes activations to a single per-layer bit width and disables DAQ; 
\noindent(ii) {No-Prune}, which restores the full timestep schedule of 1000; and
\noindent(iii) {Uniform-W8/G32}, which overwrites the learned weight plan with uniform 8 bit weights and group size 32. For reproducibility, population size, elites, generations, and mutation rates are, population $=12$, elites $=4$,  generations=$6$, mutate $\sim 10\%$ of layers per candidate. For weights, we typically use per channel/per group symmetric quantization, and for activations, we use per tensor symmetric quantization with DAQ selected precision. The results are summarized in table.~\ref{tab1}.

\subsubsection{Ablation Study Result Analysis}

From the ablation studies in table \ref{tab1}, the following insights can be gained,
i) NO-DAQ hurts quality, which means DAQ is doing actual work.
ii) No-Prune is slightly better than the full method, which denotes pruning saved time with little loss.
iii) Uniform W8 is worse than a tailored plan, which denotes per-layer bit allocation matters.

\begin{figure}[!t]
  \centering
  \includegraphics[width=0.8\columnwidth]{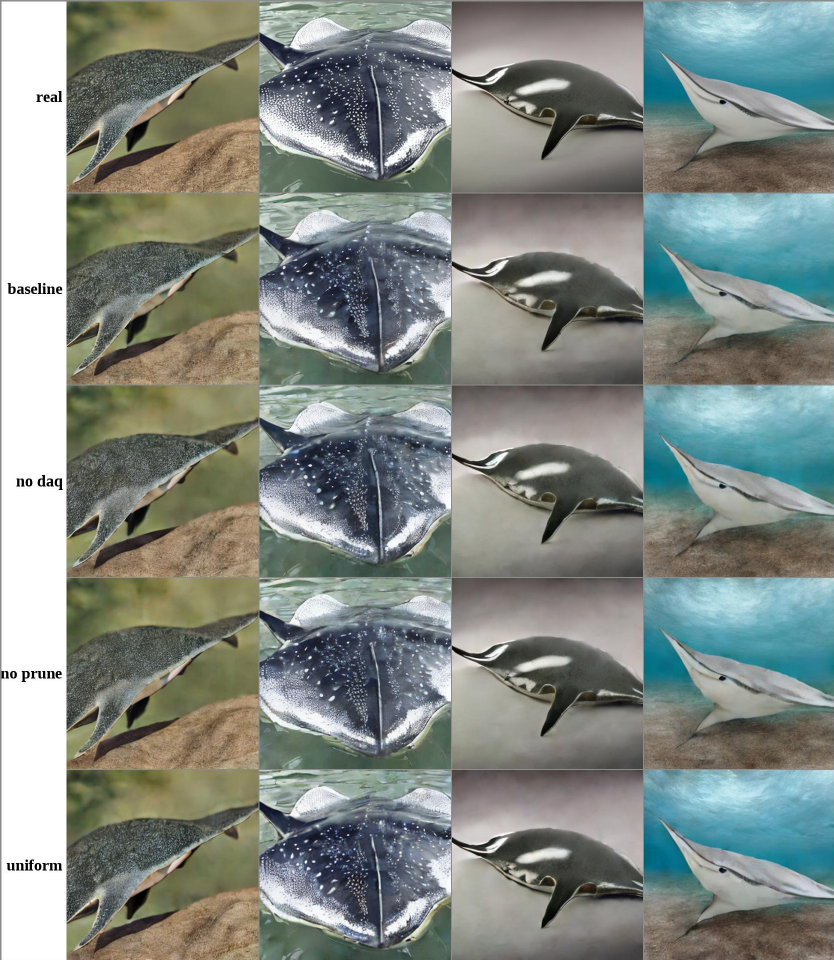}
  \caption{\textbf{Each row illustrates a generation image through ablation study. First row depicts a full precision image, second row baseline with all three components, third row without DAQ, fourth row without pruning, fifth row with a uniform 8 bit quantization.}}
  \label{fig:step_ab}
\end{figure}

Fig.~\ref{fig:step_ab} shows generated images through the ablation study.

\begin{table}[!htpb]
\caption{Main results on \emph{Imagenet} at resolution 256. Latency is per image. “Ours” uses the jointly optimized plan.}
\label{tab1}
\vskip 0.25in
\begin{center}
\begin{small}
\begin{sc}
\vspace{-1em}
\resizebox{\columnwidth}{!}{%
\begin{tabular}{lcccc}
\toprule
Method & FID $\downarrow$ & Latency(sec)  & Energy(J)& Model Size(MB)$\downarrow$ \\
\midrule
Uniform W8 / Group 32 (no DAQ)      & 31.84 & 2.72 & 389.61   & 429.949 \\
No-Prune (optimized bits + DAQ)     & 30.38 & 2.65 & 361.51   & 397.241 \\
No-DAQ (optimized bits + schedule)  & 31.17 & 2.64 & 360.64 & 403.459 \\
Full Precision  & 20 & 6.88 & 660.94 & 2575.42 \\
\textbf{All 3 (bits + DAQ + schedule)} & \textbf{28.05} & \textbf{2.55} & \textbf{360.05} & \textbf{397.241} \\
\bottomrule
\end{tabular}%
}
\end{sc}
\end{small}
\end{center}
\vskip -0.05in
\vspace{-1em}
\end{table}

\section{Conclusion}
We present DiffPro, a post training framework that (i) selects layer wise weight precision, (ii) applies lightweight per timestep DAQ, and (iii) prunes timesteps using a teacher–student drift signal and finalizes a deployable plan evaluated with real $\text{INT8}\!\times\!\text{INT8}\!\rightarrow\!\text{INT32}$ kernels. DiffPro achieves up to $6.25\times$ model compression,50\% timestep reduction with only 500 timesteps and $2.8\times$ faster inference with $\Delta\mathrm{FID}$
$\leq 10$ degradation on standard benchmarks, demonstrating practical efficiency gains. Our method uses a small calibration pass and introduces minor DAQ overhead. In  future, we plan to remove calibration needs, stabilize early steps at W4A8, improve portability across samplers, and extend DiffPro to video Diffusion Transformers.



\begin{thebibliography}{36}
\providecommand{\natexlab}[1]{#1}
\providecommand{\url}[1]{\texttt{#1}}
\expandafter\ifx\csname urlstyle\endcsname\relax
  \providecommand{\doi}[1]{doi: #1}\else
  \providecommand{\doi}{doi: \begingroup \urlstyle{rm}\Url}\fi

\bibitem[Abouelnaga et~al.(2016)Abouelnaga, Ali, Rady, and Moustafa]{abouelnaga2016cifar}
Abouelnaga, Y., Ali, O.~S., Rady, H., and Moustafa, M.
\newblock Cifar-10: Knn-based ensemble of classifiers.
\newblock In \emph{2016 International Conference on Computational Science and Computational Intelligence (CSCI)}, pp.\  1192--1195. IEEE, 2016.

\bibitem[Bao et~al.(2023)Bao, Nie, Xue, Li, Pu, Wang, Yue, Cao, Su, and Zhu]{bao2023one}
Bao, F., Nie, S., Xue, K., Li, C., Pu, S., Wang, Y., Yue, G., Cao, Y., Su, H., and Zhu, J.
\newblock One transformer fits all distributions in multi-modal diffusion at scale.
\newblock In \emph{International Conference on Machine Learning}, pp.\  1692--1717. PMLR, 2023.

\bibitem[Chai et~al.(2025)Chai, Lin, Gao, Yu, and Xie]{Chai2025Diffusion}
Chai, Z., Lin, Y.-L., Gao, Z., Yu, X., and Xie, Z.
\newblock Diffusion model empowered efficient data distillation method for cloud-edge collaboration.
\newblock \emph{IEEE Transactions on Cognitive Communications and Networking}, 11:\penalty0 902--913, 2025.
\newblock \doi{10.1109/tccn.2025.3527647}.

\bibitem[Chen et~al.(2025{\natexlab{a}})Chen, Meng, Tang, Ma, Jiang, Wang, Wang, and Zhu]{chen2025}
Chen, L., Meng, Y., Tang, C., Ma, X., Jiang, J., Wang, X., Wang, Z., and Zhu, W.
\newblock Q-dit: Accurate post-training quantization for diffusion transformers.
\newblock In \emph{Proceedings of the Computer Vision and Pattern Recognition Conference}, pp.\  28306--28315, 2025{\natexlab{a}}.

\bibitem[Chen et~al.(2025{\natexlab{b}})Chen, Mills, and Niu]{FP4DiT2025}
Chen, R., Mills, K.~G., and Niu, D.
\newblock Fp4dit: Towards effective floating point quantization for diffusion transformers.
\newblock \emph{arXiv preprint arXiv:2503.15465}, 2025{\natexlab{b}}.

\bibitem[Deng et~al.(2009)Deng, Dong, Socher, Li, Li, and Fei-Fei]{deng2009imagenet}
Deng, J., Dong, W., Socher, R., Li, L.-J., Li, K., and Fei-Fei, L.
\newblock Imagenet: A large-scale hierarchical image database.
\newblock In \emph{2009 IEEE Conference on Computer Vision and Pattern Recognition (CVPR)}, pp.\  248--255. IEEE, 2009.

\bibitem[Ding et~al.(2025)Ding, Han, Tian, Xu, Han, and Tang]{HTGDiT2025}
Ding, N., Han, J., Tian, Y., Xu, C., Han, K., and Tang, Y.
\newblock Post-training quantization for diffusion transformer via hierarchical timestep grouping.
\newblock \emph{arXiv preprint arXiv:2503.06930}, 2025.

\bibitem[Dong et~al.(2024)Dong, Zhuang, Yang, Ji, Li, Xu, Huang, Hu, Jones, Shi, Wang, and Zhou]{Dong2024EQViT}
Dong, P., Zhuang, J., Yang, Z., Ji, S., Li, Y., Xu, D., Huang, H., Hu, J., Jones, A.~K., Shi, Y., Wang, Y., and Zhou, P.
\newblock Eq-vit: Algorithm-hardware co-design for end-to-end acceleration of real-time vision transformer inference on versal acap architecture.
\newblock \emph{IEEE Transactions on Computer-Aided Design of Integrated Circuits and Systems}, 43:\penalty0 3949--3960, 2024.
\newblock \doi{10.1109/tcad.2024.3443692}.

\bibitem[Frantar et~al.(2022)Frantar, Ashkboos, Hoefler, and Alistarh]{frantar2022gptq}
Frantar, E., Ashkboos, S., Hoefler, T., and Alistarh, D.
\newblock Gptq: Accurate post-training quantization for generative pre-trained transformers.
\newblock \emph{arXiv preprint arXiv:2210.17323}, 2022.

\bibitem[Han et~al.(2023)Han, Zhao, Lv, Zhang, Liu, Bi, and Han]{Han2023Enhancing}
Han, L., Zhao, Y., Lv, H., Zhang, Y., Liu, H., Bi, G., and Han, Q.
\newblock Enhancing remote sensing image super-resolution with efficient hybrid conditional diffusion model.
\newblock \emph{Remote. Sens.}, 15:\penalty0 3452, 2023.
\newblock \doi{10.3390/rs15133452}.

\bibitem[He et~al.(2023{\natexlab{a}})He, Liu, Wu, Zhou, and Zhuang]{he2023efficientdm}
He, Y., Liu, J., Wu, W., Zhou, H., and Zhuang, B.
\newblock Efficientdm: Efficient quantization-aware fine-tuning of low-bit diffusion models.
\newblock \emph{arXiv preprint arXiv:2310.03270}, 2023{\natexlab{a}}.

\bibitem[He et~al.(2023{\natexlab{b}})He, Liu, Liu, Wu, Zhou, and Zhuang]{he2023ptqd}
He, Y., Liu, L., Liu, J., Wu, W., Zhou, H., and Zhuang, B.
\newblock Ptqd: Accurate post-training quantization for diffusion models.
\newblock \emph{arXiv preprint arXiv:2305.10657}, 2023{\natexlab{b}}.

\bibitem[Ho et~al.(2020)Ho, Jain, and Abbeel]{ho2020denoising}
Ho, J., Jain, A., and Abbeel, P.
\newblock Denoising diffusion probabilistic models.
\newblock \emph{Advances in neural information processing systems}, 33:\penalty0 6840--6851, 2020.

\bibitem[Jayawant et~al.(2025)Jayawant, Vareli, Pepper, Pepper, Simoes, and Mitchell]{Jayawant2025Computational}
Jayawant, E., Vareli, A., Pepper, A. G.~S., Pepper, C., Simoes, F., and Mitchell, S.
\newblock Computational modelling of aggressive b-cell lymphoma.
\newblock \emph{Biochemical Society Transactions}, 2025.
\newblock \doi{10.1042/bst20253039}.

\bibitem[Karras et~al.(2018)Karras, Aila, Laine, and Lehtinen]{karras2018celebAHQ}
Karras, T., Aila, T., Laine, S., and Lehtinen, J.
\newblock Progressive growing of {GANs} for improved quality, stability, and variation.
\newblock In \emph{International Conference on Learning Representations (ICLR)}, 2018.
\newblock Introduces CelebA-HQ; arXiv:1710.10196.

\bibitem[Krizhevsky et~al.(2009)Krizhevsky, Nair, and Hinton]{krizhevsky_cifar_2009}
Krizhevsky, A., Nair, V., and Hinton, G.
\newblock {CIFAR}-10 and {CIFAR}-100 datasets, 2009.
\newblock URL \url{https://www.cs.toronto.edu/~kriz/cifar.html}.
\newblock Dataset.

\bibitem[Li et~al.(2023)Li, Liu, Lian, Yang, Dong, Kang, Zhang, and Keutzer]{li2023q}
Li, X., Liu, Y., Lian, L., Yang, H., Dong, Z., Kang, D., Zhang, S., and Keutzer, K.
\newblock Q-diffusion: Quantizing diffusion models.
\newblock In \emph{Proceedings of the IEEE/CVF International Conference on Computer Vision}, pp.\  17535--17545, 2023.

\bibitem[Mills et~al.(2025)Mills, Salameh, Chen, Hassanpour, Lu, and Niu]{mills2025qua2sedimo}
Mills, K.~G., Salameh, M., Chen, R., Hassanpour, N., Lu, W., and Niu, D.
\newblock Qua2sedimo: Quantifiable quantization sensitivity of diffusion models.
\newblock In \emph{Proceedings of the AAAI Conference on Artificial Intelligence}, volume~39, pp.\  6153--6163, 2025.

\bibitem[Motetti et~al.(2024)Motetti, Risso, Burrello, Macii, Poncino, and Pagliari]{Motetti2024Joint}
Motetti, B.~A., Risso, M., Burrello, A., Macii, E., Poncino, M., and Pagliari, D.~J.
\newblock Joint pruning and channel-wise mixed-precision quantization for efficient deep neural networks.
\newblock \emph{IEEE Transactions on Computers}, 73:\penalty0 2619--2633, 2024.
\newblock \doi{10.1109/tc.2024.3449084}.

\bibitem[Pandey et~al.(2022)Pandey, Mukherjee, Rai, and Kumar]{pandey2022diffusevae}
Pandey, K., Mukherjee, A., Rai, P., and Kumar, A.
\newblock Diffusevae: Efficient, controllable and high-fidelity generation from low-dimensional latents.
\newblock \emph{arXiv preprint arXiv:2201.00308}, 2022.

\bibitem[Peebles \& Xie(2023)Peebles and Xie]{peebles2023scalable}
Peebles, W. and Xie, S.
\newblock Scalable diffusion models with transformers.
\newblock In \emph{Proceedings of the IEEE/CVF international conference on computer vision}, pp.\  4195--4205, 2023.

\bibitem[Pinheiro~Cinelli et~al.(2021)Pinheiro~Cinelli, Ara{\'u}jo~Marins, Barros~da Silva, and Lima~Netto]{pinheiro2021variational}
Pinheiro~Cinelli, L., Ara{\'u}jo~Marins, M., Barros~da Silva, E.~A., and Lima~Netto, S.
\newblock Variational autoencoder.
\newblock In \emph{Variational methods for machine learning with applications to deep networks}, pp.\  111--149. Springer, 2021.

\bibitem[Podell et~al.(2023)Podell, English, Lacey, Blattmann, Dockhorn, M{\"u}ller, Penna, and Rombach]{podell2023sdxl}
Podell, D., English, Z., Lacey, K., Blattmann, A., Dockhorn, T., M{\"u}ller, J., Penna, J., and Rombach, R.
\newblock Sdxl: Improving latent diffusion models for high-resolution image synthesis.
\newblock \emph{arXiv preprint arXiv:2307.01952}, 2023.

\bibitem[Ronneberger et~al.(2015)Ronneberger, Fischer, and Brox]{ronneberger2015u}
Ronneberger, O., Fischer, P., and Brox, T.
\newblock U-net: Convolutional networks for biomedical image segmentation.
\newblock In \emph{International Conference on Medical image computing and computer-assisted intervention}, pp.\  234--241. Springer, 2015.

\bibitem[Shang et~al.(2023)Shang, Yuan, Xie, Wu, and Yan]{shang2023post}
Shang, Y., Yuan, Z., Xie, B., Wu, B., and Yan, Y.
\newblock Post-training quantization on diffusion models.
\newblock In \emph{Proceedings of the IEEE/CVF conference on computer vision and pattern recognition}, pp.\  1972--1981, 2023.

\bibitem[Shao et~al.(2025)Shao, Lin, Zeng, Yan, Zhang, Chen, Fan, Yan, Wang, Guo, et~al.]{TRDQ2025}
Shao, Y., Lin, D., Zeng, F., Yan, M., Zhang, M., Chen, S., Fan, Y., Yan, Z., Wang, H., Guo, J., et~al.
\newblock Tr-dq: Time-rotation diffusion quantization.
\newblock \emph{arXiv preprint arXiv:2503.06564}, 2025.

\bibitem[Smith(1995)]{smith1995differentiation}
Smith, S.~P.
\newblock Differentiation of the cholesky algorithm.
\newblock \emph{Journal of Computational and Graphical Statistics}, 4\penalty0 (2):\penalty0 134--147, 1995.

\bibitem[Song et~al.(2020)Song, Meng, and Ermon]{song2020denoising}
Song, J., Meng, C., and Ermon, S.
\newblock Denoising diffusion implicit models.
\newblock \emph{arXiv preprint arXiv:2010.02502}, 2020.

\bibitem[Sun et~al.(2024)Sun, Tang, Wang, Meng, Jiang, Ma, and Zhu]{sun2024tmpq}
Sun, H., Tang, C., Wang, Z., Meng, Y., Jiang, J., Ma, X., and Zhu, W.
\newblock Tmpq-dm: Joint timestep reduction and quantization precision selection for efficient diffusion models.
\newblock \emph{CoRR}, 2024.

\bibitem[Vora et~al.(2025)Vora, Krishnan, Bouacida, Shankar, and Mohapatra]{vora2025ptq4adm}
Vora, J., Krishnan, A., Bouacida, N., Shankar, P.~R., and Mohapatra, P.
\newblock Ptq4adm: Post-training quantization for efficient text conditional audio diffusion models.
\newblock In \emph{ICASSP 2025-2025 IEEE International Conference on Acoustics, Speech and Signal Processing (ICASSP)}, pp.\  1--5. IEEE, 2025.

\bibitem[Wang et~al.(2024)Wang, Wang, Xu, Tang, Zhou, and Lu]{wang2024towards}
Wang, C., Wang, Z., Xu, X., Tang, Y., Zhou, J., and Lu, J.
\newblock Towards accurate post-training quantization for diffusion models.
\newblock In \emph{Proceedings of the IEEE/CVF Conference on Computer Vision and Pattern Recognition}, pp.\  16026--16035, 2024.

\bibitem[Wu et~al.(2024)Wu, Wang, Shang, Shah, and Yan]{wu2024ptq4dit}
Wu, J., Wang, H., Shang, Y., Shah, M., and Yan, Y.
\newblock Ptq4dit: Post-training quantization for diffusion transformers.
\newblock \emph{Advances in neural information processing systems}, 37:\penalty0 62732--62755, 2024.

\bibitem[Ye et~al.(2025)Ye, Wang, and Jiang]{ye2025pqd}
Ye, J., Wang, Z., and Jiang, L.
\newblock Pqd: Post-training quantization for efficient diffusion models.
\newblock In \emph{Proceedings of the Winter Conference on Applications of Computer Vision}, pp.\  150--156, 2025.

\bibitem[Yu et~al.(2021)Yu, Zhang, and Deng]{yu2021frechet}
Yu, Y., Zhang, W., and Deng, Y.
\newblock Frechet inception distance (fid) for evaluating gans.
\newblock \emph{China University of Mining Technology Beijing Graduate School}, 3\penalty0 (11), 2021.

\bibitem[Zhao et~al.(2024)Zhao, Fang, Huang, Liu, Wan, Soedarmadji, Li, Lin, Dai, Yan, et~al.]{ViDiTQ2025}
Zhao, T., Fang, T., Huang, H., Liu, E., Wan, R., Soedarmadji, W., Li, S., Lin, Z., Dai, G., Yan, S., et~al.
\newblock Vidit-q: Efficient and accurate quantization of diffusion transformers for image and video generation.
\newblock \emph{arXiv preprint arXiv:2406.02540}, 2024.

\bibitem[Zhou et~al.(2024)Zhou, Chen, Wang, Chen, and Lyu]{zhou2024simple}
Zhou, Z., Chen, D., Wang, C., Chen, C., and Lyu, S.
\newblock Simple and fast distillation of diffusion models.
\newblock \emph{Advances in Neural Information Processing Systems}, 37:\penalty0 40831--40860, 2024.

\end{thebibliography}

%

\end{document}